\documentclass[runningheads]{llncs}
\pdfoutput=1
\usepackage{graphicx}
\usepackage{amsmath,amssymb} 
\usepackage{color}
\usepackage{verbatim}
\usepackage{subfig}
\usepackage{array}
\usepackage{multirow}

\newcolumntype{M}[1]{>{\centering\arraybackslash}m{#1}}

\begin{document}

\title{An Improved Learning Framework for Covariant Local Feature Detection} 
\titlerunning{Improved Covariant Detection} 


\author{Nehal Doiphode\inst{1}\thanks{Work done as student in IIT Bombay} \and
Rahul Mitra\inst{2} \and
Shuaib Ahmed\inst{3} \and Arjun Jain\inst{2}}
%

\authorrunning{Doiphode et al.} 


\institute{University of Pennsylvania, USA\\
\email{lahen@seas.upenn.edu} \and
Indian Institute of Technology Bombay, India\\
\email{\{rmitter,ajain\}@cse.iitb.ac.in} \and
Mercedes-Benz Research and Development India Private Limited, India\\
\email{shuaib.ahmed@daimler.com}
}

\maketitle

\begin{abstract}
Learning feature detection has been largely an unexplored area when compared to handcrafted feature detection. Recent learning formulations use the covariant constraint in their loss function to learn covariant detectors. However, just learning from covariant constraint can lead to detection of unstable features. To impart further, stability detectors are trained to extract pre-determined features obtained by hand-crafted detectors. However, in the process they lose the ability to detect novel features.
In an attempt to overcome the above limitations, we propose an improved scheme by incorporating covariant constraints in form of triplets with addition to an affine covariant constraint. We show that using these additional constraints one can learn to detect novel and stable features without using pre-determined features for training. Extensive experiments show our model achieves state-of-the-art performance in repeatability score on the well known datasets such as Vgg-Affine, EF, and Webcam.

\end{abstract}

\section{Introduction}
Representing an image as a collection of local features is important in solving computer vision problems like generating point correspondences between images ~\cite{pt-corr} and subsequently used in Structure From Motion(SFM)~\cite{Sfm}, image stitching~\cite{image-stitching}, image retrieval~\cite{image-retrieval} and image registration~\cite{image-registration}. Hence, detecting local features from images which are invariant towards viewpoint and illumination changes has been actively pursued by the research community.

A \emph{good} local feature is characterized by two properties. First, the feature point should be discriminative from its immediate neighborhood, which ensures that these points can be uniquely identified which is essential in many vision algorithms. The second property is the ability of a feature point to be consistently detected in images which differs vastly in terms of geometric and photometric transformations. This property is termed as \emph{covariant} constraint in the literature~\cite{Ddet}. 
There have been comparatively fewer learning based detector compared to hand-crafted ones. Early learned detectors~\cite{Fast}, focused on learning to extract discriminative points. Recently, convolutional neural network (CNN) based methods ~\cite{Ddet,Covdet} coupled with covariant constraint as a loss term while training has been proposed. The method defined in~\cite{Ddet} with their CNN model DDET ensures extracted features are covariant to geometric changes without handling discriminativeness. However, as shown in~\cite{Covdet}, the feature points predicted by DDET are not stable due to the nature of covariant constraint loss. In order to predict stable features DDET was extended in COVDET~\cite{Covdet} by learning to pick pre-determined good features along with maintaining covariance. However, by regressing to pre-determined features extracted from a base detector, the learned model does not get an option to discover possibly more stable feature locations. It also inherits the nuisances of the base detectors.

To alleviate the issues faced by DDET and COVDET, we introduce a novel training framework which tries to combine the advantages of both DDET and COVDET. In our framework, we added additional geometric constraints in the form of loss functions to ensure stability of extracted features. In particular, translation covariance is enforced in multiple patches sharing a common feature point. We further incorporate affine covariance in our training procedure to increase stability towards affine transforms which is found in abundance in real image pairs. Extensive experiments show that our proposed model out-performs other approaches in three well known publicly available datasets in terms of \emph{repeatability} scores.

The contributions of this paper are the following:
\begin{itemize}
    \item We introduce a novel learning framework for covariant feature detection which extends the two existing frameworks by incorporating two different types of geometric constraints. This enables the detector to discover better and stable features while maintaining discriminativeness.
    \item The model trained with our proposed framework achieves state-of-the-art performance in \emph{repeatability} score on well known publicly available known benchmarks.
\end{itemize}

\section{Related Work}
\label{sec:related-work}
Detecting interest points in images have been dominated by heuristic methods. These methods identify specific visual structures, between images which have undergone transformations, consistently. The visual structures are so chosen to make detection covariant for certain transformation. Hand crafted heuristic detectors can be classified roughly into two category based on the visual structure they detect: i) points and ii) blobs. Point based detectors are covariant towards translation and rotation, examples include Harris~\cite{Harris},  Edge-Foci~\cite{Edge-focai}. Scale and affine covariant version of Harris~\cite{Harris} are proposed in ~\cite{Harris-scale} and ~\cite{Harris-affine} respectively. Blob detectors include DoG~\cite{Sift} and SURF~\cite{surf} are implicitly covariant to scale changes by the virtue of using scale-space pyramids for detection. Affine adaptation of blob detection is also proposed in~\cite{Harris-affine}. 

There are fewer learning based approaches compared to hand-crafted ones. The most common line of work involves  detecting anchor points based on existing detectors. TILDE~\cite{Tilde} is an example of such detector that learns to detect \emph{stable} DoG points between images taken from the same viewpoint but having drastic illumination differences. A point is assumed to be \emph{stable}, if it is detected consistently in most images sharing a common scene. By additionally introducing locations from those images where these \emph{stable} points were not originally detected into the training set, TILDE outperforms DoG in terms of repeatability. TaSK~\cite{Task} and LIFT~\cite{Lift} also detects anchor points based on similar strategies. The downside of such approaches is that the performance of the learned detector is dependent on the anchor detector used, i.e for certain transformations the learned detector can also reflect the poor performance of the anchor detector. Another area focuses on increasing repeatability of existing detectors. An instance being FAST-ER~\cite{Fast-er} improves repeatability of FAST~\cite{Fast} key-points by optimizing its parameters. In Quad-Network~\cite{quad-net} an unsupervised approach is proposed where patches extracted from two images are assigned a score and ranking consistency between corresponding patches helps in improving repeatability.

More recently, Lenc et. al~\cite{Ddet} proposed a Siamese CNN based method to learn covariant detectors. In this method, two patches related by a transformation are fed to the network which regresses two points (one for each patch). Applying a loss ensuring that the two regressed points differ by the same transformation, lets the network detect points covariant to that transformation. However, a major drawback of the method is the lack of ensuring the network regressing to a \emph{good} and \emph{stable} feature point. This method can lead to the CNN model being trained sub-optimally. In order to alleviate the drawback, Zhang et. al~\cite{Covdet} proposed a method using standard patches to ensure that the network regress to the keypoints. The standard patches are centered around a feature detected by detectors such as TILDEP24~\cite{Tilde}. Though, this method of standard patches is generic in nature, the transformation extensively studied is translations.

\section{Fundamentals of Covariant Detection}
\label{sec:proposed-method}
\subsection{Preliminary}
\label{sec:prelims}

\label{sec:covariant-fundamentals}
Let $\mathbf{x}$ be an image patch from a set of all patches $\mathcal{X}$. A feature $\mathbf{f} \in \mathcal{F}$ is represented by a point (position w.r.t center of the patch) or a circle (position and scale) or an ellipse (position, scale, shape). Let $\mathbf{g}$ be a function representing geometric transformation belongs to a group of transformation $G$. Let $\mathbf{f_0}$ be a fixed canonical feature positioned at the center point of the patch and unit scale. When $\mathcal{F}$ resolves $\mathcal{G}$, there will be a bijective mapping between $\mathcal{F}$ and $\mathcal{G}$~\cite{Ddet,Covdet}
With this setting, instead of feature detector, it is possible to work with transformations and  a function $\phi(\cdot)$ usually represented by a CNN can take a patch $\mathbf{x}$ as input and regresses a transformation $\mathbf{g}$ that brings a \emph{good} feature positioned at $\mathbf{f}$ to $\mathbf{f_0}$. In other words, $\phi(\mathbf{x}) \:=\: \mathbf{g}; \: \mathbf{f_0} \:=\: \mathbf{g} \otimes \mathbf{f}$,  $\otimes$ is an operator used to transform $\mathbf{f}$ using $\mathbf{g}$. Now, the function $\phi()$ respects covariant constraints for a class of transformations $G$ when the Eq.~\ref{cov-constraint} holds for all patches and all transformations belonging to the class.
\begin{equation}
\label{cov-constraint}
\phi(\mathbf{g} * \mathbf{x}) \: = \: \mathbf{g} \circ \phi(\mathbf{x}); \: \forall \mathbf{x} \in \mathcal{X}, \forall \mathbf{g} \in G
\end{equation}

In the above equation $*$ symbolizes warping of patch $\mathbf{x}$ with transformation $\mathbf{g}$ and $\circ$ stands for composition of transformations. 
\subsection{Covariant constraint}
In Lenc. et al~\cite{Ddet} a Siamese network takes two patches $\mathbf{x}$(reference patch) and $\mathbf{x^{\prime}} \:=\: \mathbf{g} * \mathbf{x}$ and uses the covariant constraint mentioned in Eq.~\ref{cov-constraint} as the loss function. The optimum regressor $\phi(\cdot)$ is obtained as,
\begin{equation}
\phi \:=\:  \operatorname*{argmin}_\phi \sum_{i=1}^{n} {\lVert \phi(\mathbf{g_i} * \mathbf{x_i}) - \mathbf{g_i} \circ \phi(\mathbf{x_i}) \rVert}^2.
\end{equation}

However, just using the covariant constraint makes the learning objective have multiple solutions as mentioned in ~\cite{Covdet}. This can makes learning ambiguous and the can lead the regressor to choose non-discriminative points. 
\subsection{Standard patches}

In order to avoid the above mentioned limitations of only using covariant constraints, Zhang et. al.~\cite{Covdet} introduced the concept of \emph{standard patches} and shown that the function $\phi(\cdot)$ trained only with such patches is sufficient to mitigate the limitation. The standard patches $\mathbf{x}$ are  reference patches centered around any feature detectors such as TILDEP24~\cite{Tilde} points. An additional loss term(\emph{identity loss}) is incorporated which ensures $\phi(\cdot)$ regresses to the center point in $\mathbf{x}$. The optimization objective including both the covariant loss and the identity loss is shown in Eq.~\ref{covdet-loss}.
\begin{equation}
\label{covdet-loss}
\phi \:=\:  \operatorname*{argmin}_\phi \sum_{i=1}^{n} {\lVert \phi(\mathbf{g_i} * \mathbf{x_i}) - \mathbf{g_i} \circ \phi(\mathbf{x_i}) \rVert}^2 + \alpha \sum_{i=1}^{n} {\lVert \phi(\mathbf{x_i}) \rVert}^2
\end{equation}

\begin{figure}
    \centering
    \includegraphics[width=0.8\linewidth]{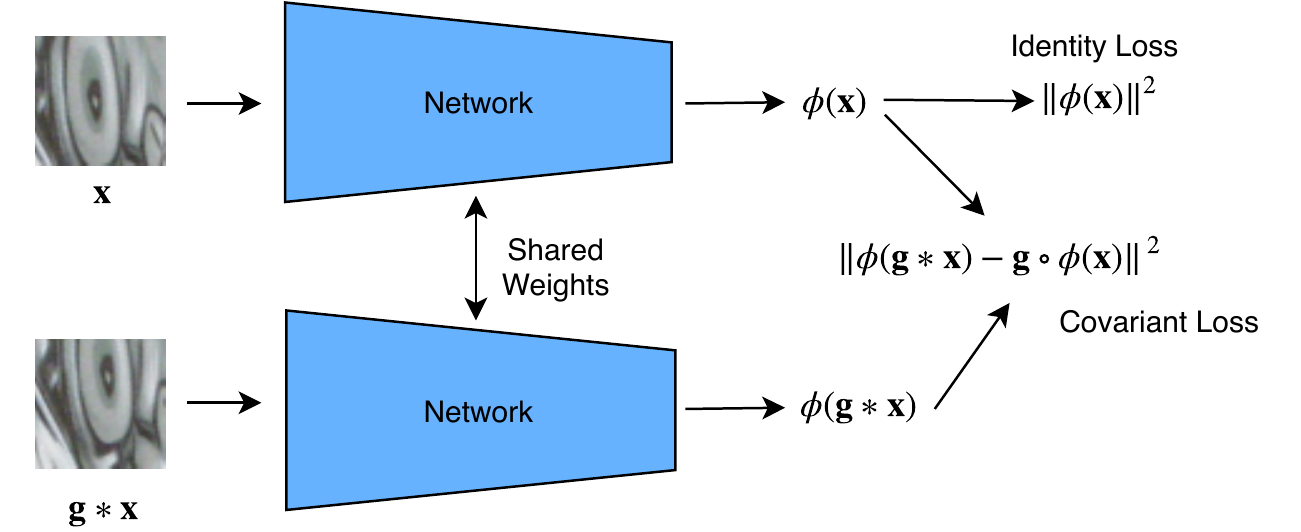}
    \caption{The Siamese architecture along with the \emph{covariant} and \emph{identity} loss used in~\cite{Covdet}.}
    \label{covdet-net}
\end{figure}
\begin{figure}
    \centering
    \includegraphics[scale=0.6]{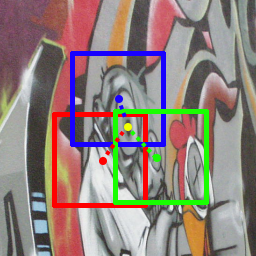}
    \caption{The 3 adjacent patches outlined in red, blue and green sharing a common good feature point marked in yellow. The dotted lines are predicted translations of the feature point from the center of their respective patch.}
    \label{common-feat}
\end{figure}

The network architecture used in~\cite{Covdet} along with the loss terms is shown in Fig.~\ref{covdet-net}. Both DDET~\cite{Ddet} and COVDET~\cite{Covdet} limits the class of transformations to translations. This implicitly makes the network to regress to same feature point which is shared among adjacent patches as shown in Fig.~\ref{common-feat}.

\begin{figure}
    \centering
    \subfloat[][Predicted 2D translations]{
    \includegraphics[width=0.4\textwidth]{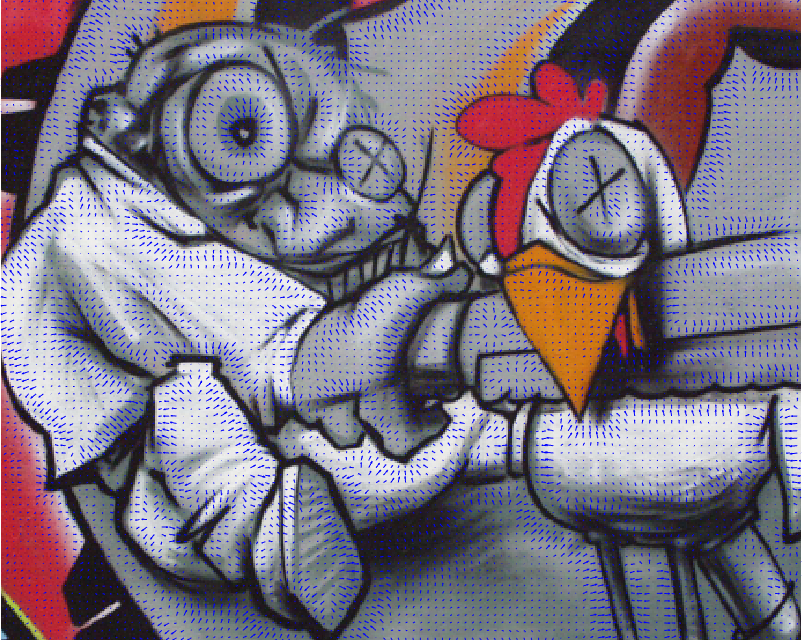}}
    ~
    \subfloat[][Generated vote-map]{
    \includegraphics[width=0.4\textwidth]{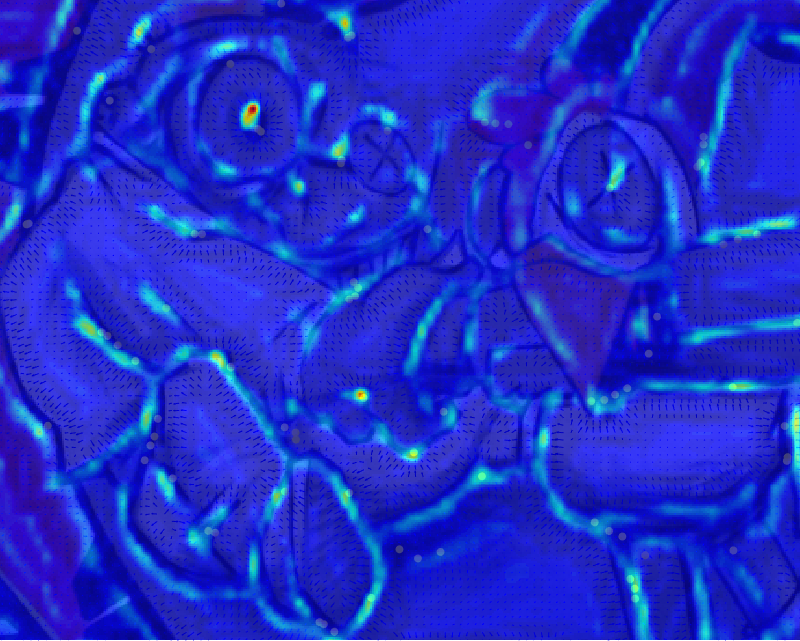}}
    
    \subfloat[][Final extracted points]{
     \includegraphics[width=0.4\textwidth]{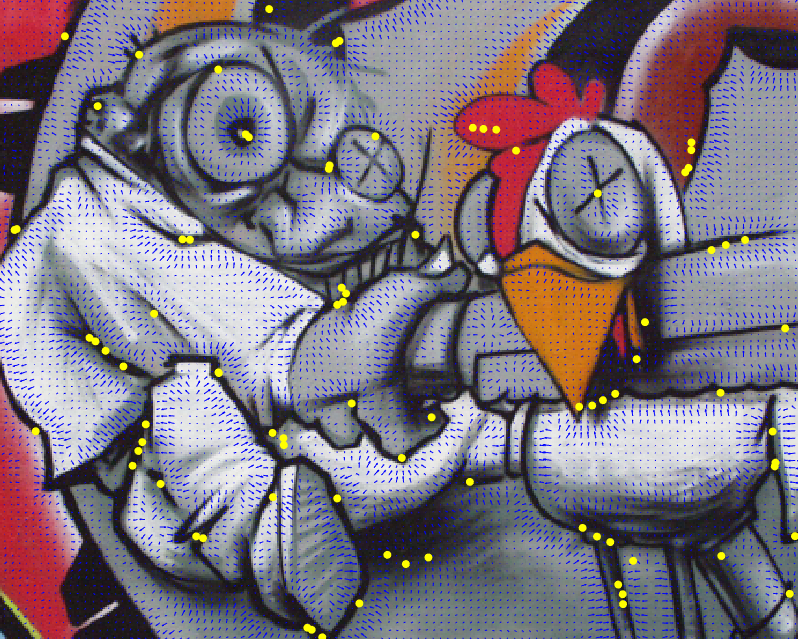}}
    
    \caption{Output of different stages of feature extraction pipeline for a full image. (a) Shows the predicted 2D translations for every $4^{th}$ pixel. (b) Shows the bi-linearly interpolated votemap (c) Shows the final detected features.}
    \label{image-extract}
\end{figure}

While extracting features from an image, the entire image is forwarded through the network. The network being fully-convolutional, outputs the translations to feature points relative to 2D grid of every 4 pixels (since both DDET and COVDET has two $2 \times 2$ pool layers) taken as centers. The predicted translations are added to their respective center pixels to get final positions of the regressed points. A vote-map is generated by bi-linear interpolation of the regressed locations to its 4 nearest pixels and accumulating the contributions. Since, grid points sharing a good feature point regress to its location, the vote-map exhibits high density at locations of these points. Finally, similar to many feature point detection algorithms, non-maximal suppression is used to select points with local maximum. In Fig.~\ref{image-extract}(a) and (b) predicted translations for each grid point and the generated vote-map is shown respectively. Fig.~\ref{image-extract}(c) shows the final selected points after non-maximal suppression. The usage of standard patches improves the repeatability and matching score of the learned detector when compared to DDET~\cite{Ddet}.

\section{Proposed Method}
\label{sec:Proposed-Method}
The method described in ~\cite{Ddet} and its extension~\cite{Covdet} suffer from a few drawbacks as follows:
\begin{itemize}
    \item Using only a covariance loss between the reference patch and translated patch can introduce instability in the learning procedure as mentioned in Sec.~\ref{sec:covariant-fundamentals}. The regressed points are also not ensured to be discriminative. 
    \item Regressing to pre-determined feature points extracted by a base detector resolves issues of instability but makes the model susceptible to the vulnerabilities of the base detector.
\end{itemize}

To overcome the drawbacks, we extend the work of~\cite{Covdet} by introducing a novel training framework which incorporates multiple neighborhood patches sharing a \emph{good} feature while designing the covariant constraint. Affine covariance between reference patch and affine warp of it is also introduced. The additional geometric constraints enforces the network to choose points which are stable to such transformation and in effect reducing the regression space. In addition to the above, the proposed model is not explicitly enforced to regress to pre-computed TILDE~\cite{Tilde} points. This is done in accordance with observations in~\cite{Lift} where the authors observed sub-optimal performance when only selecting points retrieved by SFM rather than treating the location as latent variable and letting the model discover more reliable points in the vicinity. 

The proposed training framework, in essence, similar to descriptor training frameworks L2-Net~\cite{L2-net} and Hardnet~\cite{Hardnet} where the loss functions are formulated by coupling together a batch of triplets. Their modified framework outperforms Tfeat~\cite{Tfeat} which trains using triplets independently. The reason attributed to the improvement in performance is increase in context which in case of descriptor training comes with gradients from several coupled triplets. In our case, additional geometric constraints obtained from coupling all the pairs serves as the extra context needed for stability. 

In the proposed extension, a triplet of patches $\mathbf{x_1}, \mathbf{x_2}, \mathbf{x_3}$ from a reference patch $\mathbf{x}$ is generated by translating $\mathbf{x}$ by $\mathbf{t_1}, \mathbf{t_2}, \mathbf{t_3}$ respectively. Translation covariance is enforced between any of the two patches taken in order in the learning framework. Further a fourth patch $\mathbf{x_A}$ which is an affine warp (using affine transform $\mathbf{A}$) of $\mathbf{x}$ is created to ensure the regressed point is stable towards affine transformations. It should be noted that the reference patch does not necessarily have \emph{good} feature point at the center. The advantage of our approach over DDET~\cite{Ddet} which also does not regress to pre-determined features is the enforcement of additional geometric constraints by introducing multiple covariance constraints patches, $\mathbf{x}, \mathbf{x_1}, \mathbf{x_2}, \mathbf{x_3}$ and $\mathbf{x_A}$. These additional constraints ensures selection of more stable features than DDET. 
Since, the learned regressor should obey Eq.~\ref{cov-constraint} for all the pairs formed by the reference patch and the generated patches, we have the following set of equations,
\begin{subequations}
\begin{align}
    \phi(\mathbf{x_1}) \:=\: \phi(\mathbf{x}) + \mathbf{t_1} \\
    \phi(\mathbf{x_2}) \:=\: \phi(\mathbf{x}) + \mathbf{t_2} \\
    \phi(\mathbf{x_3}) \:=\: \phi(\mathbf{x}) + \mathbf{t_3} \\
    \phi(\mathbf{x_A}) \:=\: \mathbf{A} * \phi(\mathbf{x})
    \label{trip-cov-constraint}
\end{align}
\label{trip-cov-constraint}
\end{subequations}
The covariant constraint between any two of the patches from the generated triplet along with reference patch is given in Eq.~\ref{trip-cov},
\begin{subequations}
\begin{align}
\alpha \phi(\mathbf{x_1}) - \beta \phi(\mathbf{x_2}) \:=\: (\alpha - \beta) \phi(\mathbf{x}) + \mathbf{t_{12}} \\
\alpha \phi(\mathbf{x_2}) - \beta \phi(\mathbf{x_3}) \:=\: (\alpha - \beta) \phi(\mathbf{x}) + \mathbf{t_{23}} \\
\alpha \phi(\mathbf{x_3}) - \beta \phi(\mathbf{x_1}) \:=\: (\alpha - \beta) \phi(\mathbf{x}) + \mathbf{t_{31}}
\end{align}
\label{trip-cov}
\end{subequations}
The above equation is obtained from the first 3 sub equations of Eq.~\ref{trip-cov-constraint}. $\mathbf{t_{ij}}$ in the above equations equals $\alpha \mathbf{t_i} - \beta \mathbf{t_j}$. The constants $\alpha$ and $\beta$ are chosen to be 2 and 1 respectively. Higher values of $\alpha$ and $\beta$ led to instability in training due to high gradient flows.

\begin{figure}[h]
    \centering
    \includegraphics[width=0.75\linewidth]{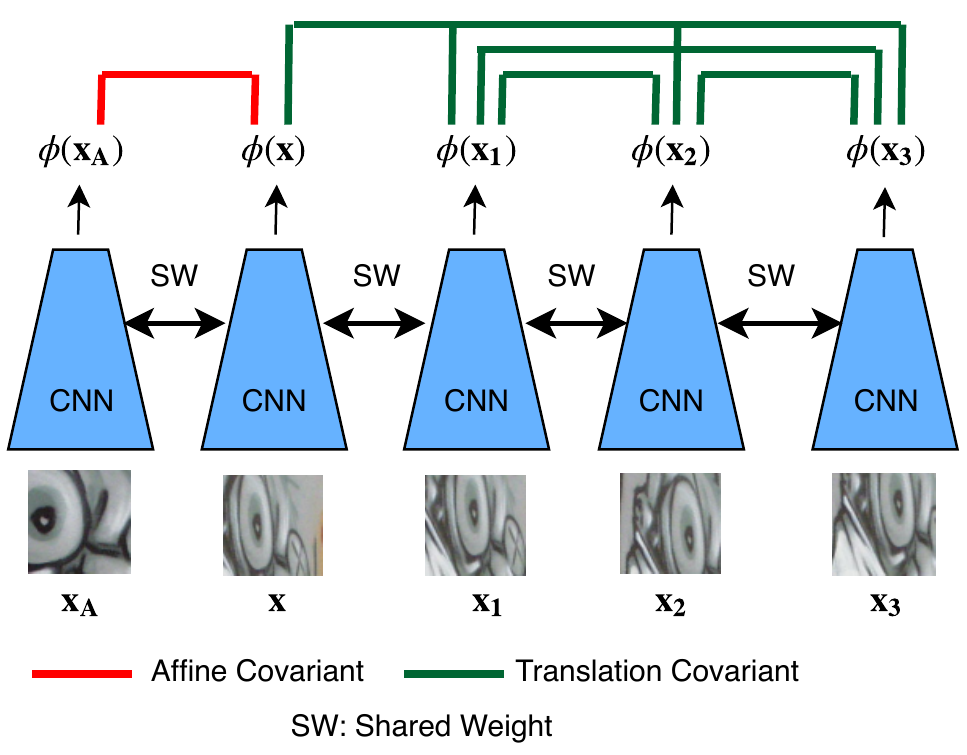}
    \caption{Training framework of our proposed approach. The green lines connect outputs which are used in the translation covariant loss terms while outputs connected by the red line contribute in the affine-covariant loss term.}
    \label{proposed-net}
\end{figure}

Our loss term has 2 components, the first component comprises of the translation covariant terms which are derived from Eq.~\ref{trip-cov} and formulated in Eq.~\ref{trans-cov-loss}.
\begin{multline}
    \ell_{cov-tran} \:=\: {\lVert \alpha \phi(\mathbf{x_1}) - \beta \phi(\mathbf{x_2}) - (\alpha - \beta) \phi(\mathbf{x}) - \mathbf{t_{12}} \rVert}^2 + \\ {\lVert \alpha \phi(\mathbf{x_2}) - \beta \phi(\mathbf{x_3}) - (\alpha - \beta) \phi(\mathbf{x}) - \mathbf{t_{23}} \rVert}^2 + \\ {\lVert \alpha \phi(\mathbf{x_3}) - \beta \phi(\mathbf{x_1}) - (\alpha - \beta) \phi(\mathbf{x}) - \mathbf{t_{31}} \rVert}^2
    \label{trans-cov-loss}
\end{multline}
In addition to the above, a second loss term ensuring affine covariance between reference patch $\mathbf{x}$ and its affine warp $\mathbf{x_A}$ is defined in Eq.~\ref{aff-cov-loss}.
\begin{equation}
    \centering
    \ell_{cov-aff} \:=\: {\lVert \phi(\mathbf{x_A}) - A * \phi(\mathbf{x}) \rVert}^2
    \label{aff-cov-loss}
\end{equation}
The total loss is given in Eq.~\ref{loss-total},
\begin{equation}
    \ell_{total} \:=\: \ell_{cov-tran} + \ell_{cov-aff}
    \label{loss-total}
\end{equation}
Our training framework is shown in Fig.~\ref{proposed-net}. It takes a tuple of 5 patches and passes each of them through a network of same configuration and sharing weights. The outputs of the triplet patches contribute to the translation covariant loss term while that of the affine warped patch to affine covariant term. The output of the reference patch is utilized in both the loss terms. We restrict to training on tuple of 5 patches as increasing the number leads to a combinatorial increase in the number of terms in the loss function which are difficult to converge during training.

\section{Experimental Setup}
\label{sec:experimental-setup}

Details of implementation and training procedure are mentioned in Sec.~\ref{sec:implementation}. Sec.~\ref{sec:eval-prot} explains the different evaluation protocols and characteristics of test-datasets used.

\subsection{Implementation And Training Details}
\label{sec:implementation}
For fair comparisons, the same network architecture and training data used in~\cite{Covdet} has been used. The input patches are of size $32 \times 32$ pixels. Table.~\ref{covdet-layers} gives the different layers of the network used in~\cite{Covdet}. In Table.~\ref{covdet-layers}, A max pool layer with kernel $2 \times 2$ is used after convolution layers 1 and 2. After each convolution layer except the last layer, ReLU activation is used. As in~\cite{Covdet} only position is regressed, hence the output of the last layer is 2.
\begin{table}[h]
\centering
\caption{The configuration of the different convolution layers used in the CNN.}
\label{covdet-layers}
\begin{tabular}{||M{2.0cm}|M{1.0cm}|M{1.0cm}|M{1.0cm}|M{1.0cm}|M{1.0cm}||}
    \hline
     layers & 1 & 2 & 3 & 4 & 5 \\
     \hline
     kernel, \#feats & 5, 32 & 5, 128 & 3, 128 & 3, 256 & 1, 2 \\
     \hline
\end{tabular}
\end{table}

For generating the patches for training, the same set of standard patches which are used in~\cite{Covdet} is used. These patches are extracted from the \emph{Mexico} scene of the Webcam Dataset~\cite{Tilde}. The same set of perturbation applied to the reference patch is maintained like scaling both axes uniformly within $[0.85, 1.15]$, shearing both axis within $[-0.15, 0.15]$, and uniform rotation between $[0^{\circ}, 360^{\circ}]$. Additionally, to ensure that the reference patch does not necessarily have a good feature point at the center, the reference patch is translated randomly by $[-5, 5]$ in both directions. To generate the triplet of patches $\mathbf{x_1}, \mathbf{x_2}, \mathbf{x_3}$ mentioned in Sec.~\ref{sec:proposed-method}, the reference patch is randomly translated between $[-6, 6]$. For affine covariance, the patch $\mathbf{x_A}$ is generated by applying another affine warp to the reference patch with the scale, shear, and rotation values sampled ranges mentioned earlier. A 256,000 samples of the tuple $\mathbf{x}, \mathbf{x_1}, \mathbf{x_2}, \mathbf{x_3}, \mathbf{x_A}$ has been generated for our training. The model is trained for 10 epochs with batch size of 128. For the first 5 epochs the affine loss term (Eq.~\ref{aff-cov-loss}) is turned off and trained with only translation covaraince loss (Eq.~\ref{trans-cov-loss}). For the next 5 epochs, both the loss terms are used. It has been observed that training with both the translation and affine loss terms from the beginning gives sub-optimal results. One possible explanation would be that the affine loss is difficult to satisfy for an untrained network. For optimization, SGD with momentum 0.9 and initial learning rate 0.1 is used. The learning rate is exponentially decayed with a decay rate of 0.96 after every epoch.
\subsection{Evaluation Protocols}
\label{sec:eval-prot}
Two publicly available datasets are used for evaluation purposes,
\begin{itemize}
    \item \textbf{VggAffine} dataset~\cite{Vgg-affine}, contains 8 scenes or sequences with 6 images per sequence. The images are numbered 0 through 5 and the image with serial 0 in the sequence is the reference image. The amount of transformation increases in the image with increase in its serial number. The transformations are related by ground truth homography. Each sequence exhibits either geometric or illumination transforms.  

    \item \textbf{Webcam} dataset~\cite{Tilde}, contains 6 sequences with 80, 20 and 20 images for training, testing and validation respectively, per sequence. All images in a sequence is taken from the same viewpoint. The images are in different times of days and at different seasons providing extreme illumination and appearance changes.

    \item \textbf{EF} dataset~\cite{Edge-focai}, contains 5 scenes or image sequences showing simultaneous variation in viewpoint, scale, and illumination. Like Hpatches, images in a sequence are related by homography.
\end{itemize}
As with COVDET~\cite{Covdet}, \emph{repeatability} and \emph{matching score} are used as the metrics for quantitative evaluation of the proposed approach and for comparison against others.
\begin{itemize}
\item \textbf{Repeatability}: It is the same evaluation procedure detailed in~\cite{Vgg-affine}. In this procedure, regions around extracted feature points are considered instead of using only the location. The size of the regions depends on the scale at which the the feature was extracted. Two regions $R_A$ and $R_B$ from images $A$ and $B$ are said to correspond when the overlap between $R_A$ and projection of $R_B$ into image $A$ is more than $40\%$. Since repeatability score for a detector is sensitive to the number of points extracted per image, all models are evaluated twice by fixing the number of points extracted to 200 and 1000.

\item \textbf{Matching score}: repeatability of a detector although useful cannot quantify the discriminative nature of the extracted points. One way to evaluate discriminativeness of the extracted points is to compute descriptors of regions around the points and use them for matching using nearest neighbors. The protocol used in Tfeat~\cite{tfeat} has been followed where in-order to eliminate any bias between certain image pairs, 1000 points are extracted from each image and SIFT~\cite{Sift} descriptors are computed. Matching score is the fraction of correct matches to the total number of points extracted. Out of any two detectors having comparable repeatability, the one with more discriminative feature points will have a better matching score.
\end{itemize}

\section{Results}
\label{sec:results}
This section discusses about the comparative results of other popular approaches against our proposed in the two evaluation criteria mentioned in Sec.~\ref{sec:eval-prot}. Table.~\ref{modif-eval} provides an ablation on the two different component of our loss terms. Additionally, a model trained using the training framework of COVDET~\cite{Covdet} but with twice the amount of training pairs is also compared. We call this model as COVDET++. Comparison with this model is done to understand the effect of training with additional data. \emph{Trip} represents the model trained using only loss mentioned in Eq.~\ref{trans-cov-loss}. \emph{Cov} + \emph{Aff} represents a model trained with loss which is a combination of one translation covariance term and the affine loss term. \emph{Trip} + \emph{Aff} is our proposed method and trained with the loss given in Eq.~\ref{loss-total}. The mean and standard deviation of repeatability score of each model from 5 training runs is reported. 1000 points were extracted from each image. As can be seen from Table.~\ref{modif-eval}, our proposed approach of combining translation covariance with affine covariance clearly outperforms other variations. The superior performance of our proposed method along with low standard deviation values validates the need to have additional geometric constraints to extract points stable towards geometric covariance.
\begin{table}[h]
\centering
\caption{The mean $\pm$ std.dev of repeatability on all datasets for various modifications of proposed approach.}
\label{modif-eval}
\begin{tabular}{||l|M{2.5cm}|M{2.5cm}|M{2.5cm}||}
\hline
Method & Vgg-Aff & EF & Webcam\\ \hline\hline 
COVDET++     &  $58.14\pm0.12$ & $36.41\pm0.05$ & $51.39\pm0.05$\\ \hline
\emph{Trip}  &  $54.65\pm1.34$ &  $32.10\pm0.30$ &  $48.45\pm0.30$\\ \hline
\emph{Cov} + \emph{Aff} &  $59.72\pm1.15$ &  $34.14\pm0.20$  & $50.23\pm0.30$\\ \hline
\emph{Trip} + \emph{Aff}(proposed)  &  $\mathbf{63.12}\pm0.20$  &  $\mathbf{36.50}\pm0.05$ &  $\mathbf{51.40}\pm0.05$\\ \hline
\end{tabular}
\end{table}

Table.~\ref{comp-rep} provides a comparative measure of repeatability on all the 3 datasets mentioned in Sec.~\ref{sec:experimental-setup} for our proposed approach against other popular approaches. Our proposed approach either outperforms or performs comparably on the all the three benchmark datasets. The performance gap is higher in case of 200 points than for 1000 points which in line with the observations of COVDET~\cite{Covdet}.  
\begin{table}
\centering
\caption{Comparing repeatability \% of different methods on all the datasets. 200 and 1000 represents the number of features extracted per image.}
\label{comp-rep}
\begin{tabular}{||l|M{1.4cm}|M{1.4cm}|M{1.4cm}|M{1.4cm}|M{1.4cm}|M{1.4cm}||}
\hline
\multirow{2}{*}{Method} & \multicolumn{2}{c|}{Vgg-Affine} & \multicolumn{2}{c|}{EF} & \multicolumn{2}{c||}{Webcam}\\
\cline{2-7}
                           & 200      & 1000    & 200    & 1000    & 200    & 1000 \\ \hline\hline
TILDEP24~\cite{Tilde}     & 57.57    &  64.35  & 32.3  &  45.37   & 45.1   & 61.7 \\ \hline
DDET~\cite{Ddet}           &  50.9    & 65.41   & 24.54  &  43.31  & 34.24  & 50.67 \\ \hline
COVDET~\cite{Covdet}       &  59.14   & 68.15   & 35.1   &  46.10  & 50.65  & \textbf{67.12} \\ \hline
Proposed                       &  \textbf{63.12}   & \textbf{69.79} &  \textbf{36.5}   & \textbf{46.49}         &  \textbf{51.4}  &   65.0259    \\ \hline
\end{tabular}
\end{table}

As discussed in Sec.~\ref{sec:experimental-setup}, repeatability does not takes into account the discriminitiveness of the extracted feature points. Matching feature points using a feature descriptor gives an estimate of feature discriminativeness. Table.~\ref{match-score} reports the matching score coupling SIFT~\cite{Sift} as feature descriptor with different feature extractors. We observe that our proposed model clearly outperforms DDET~\cite{Ddet}, while performing comparably or marginally below COVDET~\cite{Covdet}. These results translates that the feature points detected by the proposed models are better discriminatable than DDET and comparable to COVDET. 
\begin{table}
\centering
\caption{Comparing Matching Score of different methods on all the datasets}
\label{match-score}
\begin{tabular}{||l|M{2.0cm}|M{2.0cm}|M{2.0cm}|M{2.0cm}||}
\hline
Method & Vgg-Aff & EF & Webcam & Average\\ \hline

TILDEP24~\cite{Tilde}     &   36.13  &  5.50 &  13.41     & 18.34       \\ \hline
DDET~\cite{Ddet}           &   37.09  &  4.15 &  12.56    &     17.93  \\ \hline
COVDET~\cite{Covdet}       &   41.79  &  \textbf{6.15} & 19.19     &  22.37     \\ \hline
Proposed                       &   \textbf{41.81}  &  5.85 & \textbf{20.14}   & \textbf{22.60}      \\ \hline
\end{tabular}
\end{table}

\section{Qualitative Analysis}
\label{sec:discussion}
In this section, visual differences between extracted features of our proposed model with that of COVDET and TILDE is analyzed. 
\begin{figure}
    \centering
    \subfloat[][Extracted feature points from `img5` of \emph{Wall}. Left: proposed, Right: COVDET]{
    \includegraphics[width=0.75\textwidth]{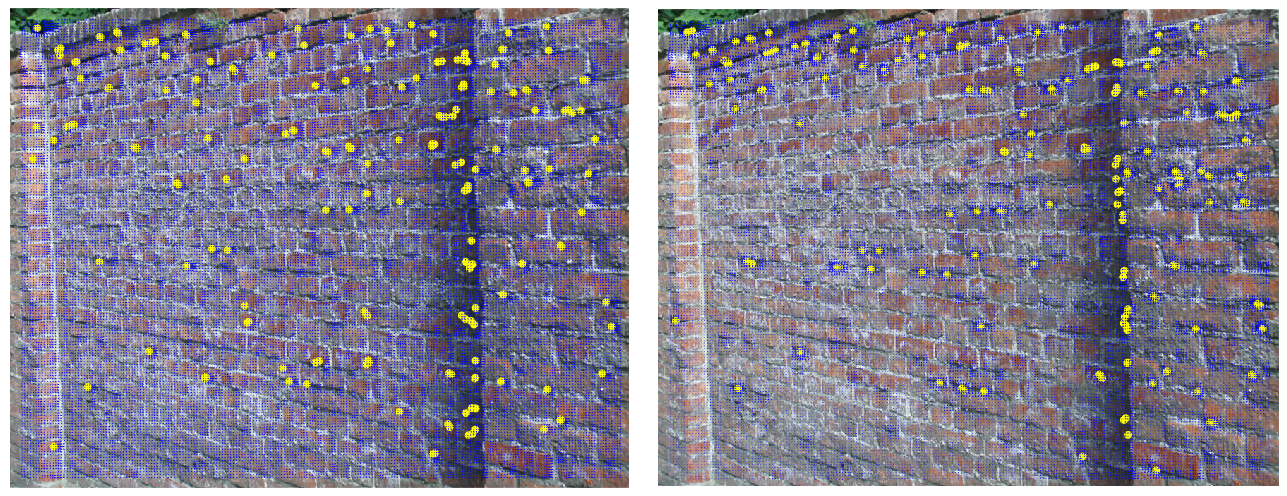}}

    \subfloat[][Extracted feature points from `img6` of \emph{Wall}. Left: proposed, Right: COVDET]{

    \includegraphics[width=0.75\textwidth]{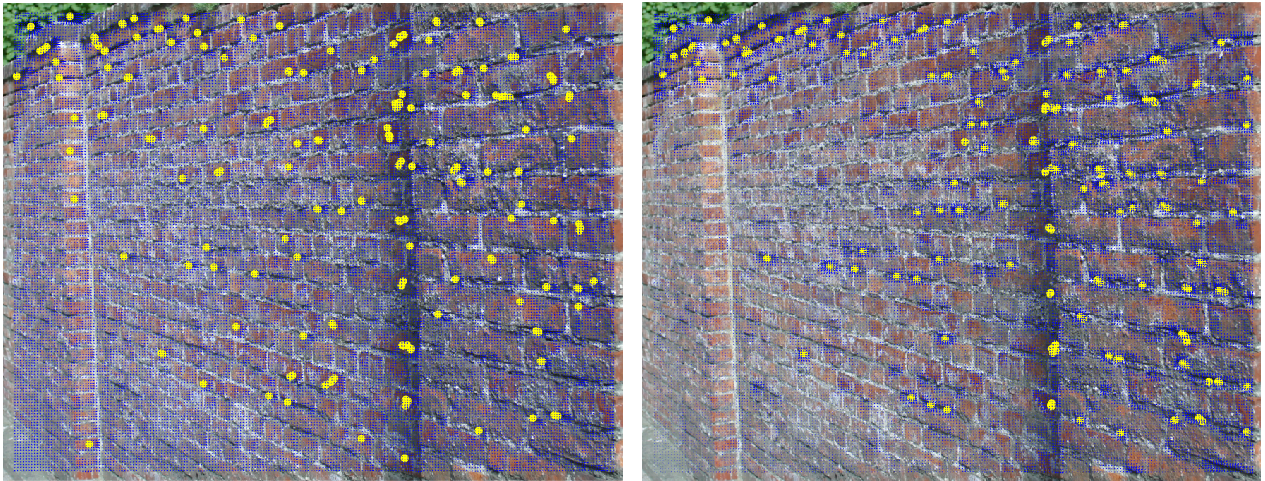}}

    \subfloat[][Extracted feature points from `img6` of \emph{Yosemite}. Left: proposed, Right: COVDET]{

    \includegraphics[width=0.75\textwidth]{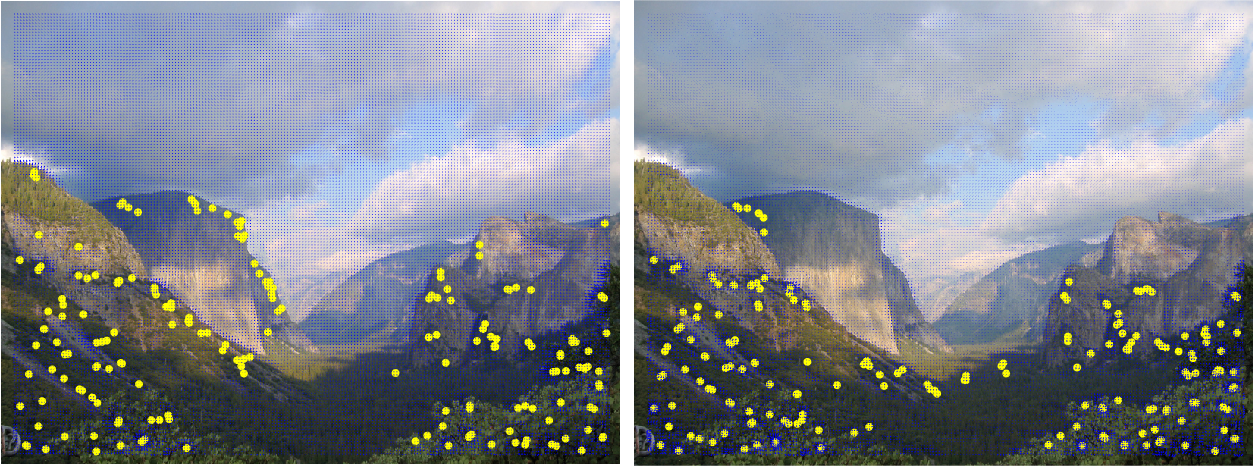}}

    \caption{Features extracted from the \emph{Wall} sequence from Vgg-Affine~\cite{Vgg-affine} and \emph{Yosemite} sequence from  EF~\cite{Edge-focai} dataset.}
    \label{comp-columbia-our-pts}
\end{figure}

A visual comparison between extracted feature from images of \emph{Wall} and \emph{Yosemite} sequence is shown Fig.~\ref{comp-columbia-our-pts}. It can be observed that our proposed method have extracted features which are far more spread out in the images than the ones by COVDET. Also, from visual inspection most of the points extracted by our model lies inside discriminative areas.  

\begin{figure}
    \centering
    \includegraphics[width=0.75\textwidth]{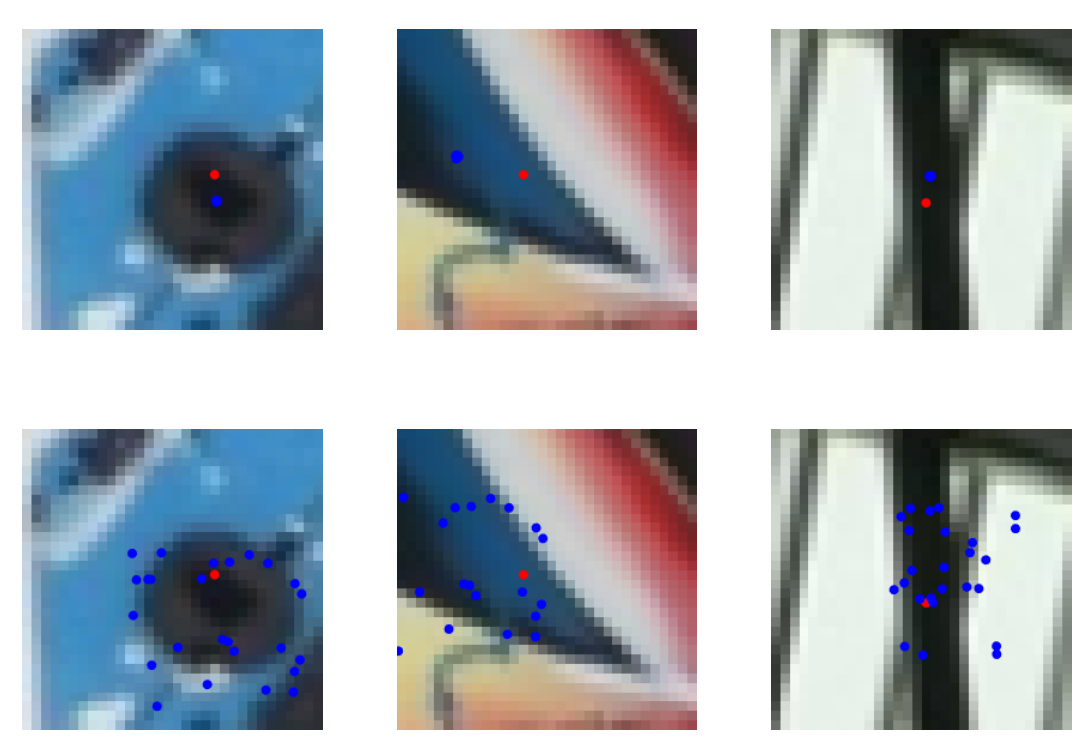}
    \caption{The top row shows 25 points predicted by the model trained with affine covariance loss while the bottom shows the predictions for the model trained without affine covariance loss. The points marked in red is a pre-determined feature point. The points marked in blue are predicted by the model.}
    \label{scatter-plot}
\end{figure}

In order to visualize the impact of using the affine covariance loss, two models were trained, which regresses 25 points instead of 1, for a given patch. Both the models are trained using the \emph{Mexico} scene of the \emph{Webcam} dataset. Covariant losses are applied individually to each of the 25 points. The only difference between the two models is that, one is trained with additional affine covariance and the other one without. Fig.~\ref{scatter-plot} investigates the distribution of the points regressed by the two models on unknown test patches containing a predetermined feature point. We can see that all the points regressed by the model trained with the affine loss collapses into a single point close to the feature point there by enforcing the stability of that point. We can see quite the opposite for the model trained without affine loss with point scattered around.

\section{Conclusion}
The existing learned covariant detectors have the drawbacks of detecting either unstable features or constraint themselves not to detect novel features. In this work, we have addressed these limitations by proposing an approach that uses covariant constraints in a triplet fashion and also incorporates an affine constraint for robust learning of the keypoint detector. Multiple neighborhood patches sharing a \emph{good} feature has been incorporated, while designing the covariant constraint. As a result, the proposed network learned novel features on its own further resulting in an improvement in the repeatability measure on state-of-the-art benchmarks. Lower standard deviation in the repeatability score for multiple runs validates the stability of the features detected. From these observation, we conclude that the proposed method including affine loss term guides the regressor in choosing a stable and novel features. Future approaches may include to extract both shape and position information from a single network.


%
%
%
\bibliographystyle{splncs04}
%

\end{document}